\pdfoutput=1

\documentclass[11pt]{article}

\usepackage[preprint]{coling}

\usepackage{times}
\usepackage{latexsym}

\usepackage[T1]{fontenc}

\usepackage[utf8]{inputenc}

\usepackage{microtype}

\usepackage{inconsolata}

\usepackage{graphicx}

\usepackage{multirow}
\usepackage{booktabs}
\usepackage{tabularx}
\usepackage{amsmath}
\usepackage{amsfonts}
\usepackage{makecell}
\usepackage{xcolor}
\usepackage{pifont}
\usepackage{comment}
\usepackage{subcaption}

\title{DeltaDQ: Ultra-High Delta Compression for Fine-Tuned LLMs via Group-wise Dropout and Separate Quantization}

\author{
 \textbf{Yanfeng Jiang\textsuperscript{1}},
 \textbf{Zelan Yang\textsuperscript{2}},
 \textbf{Bohua Chen\textsuperscript{2}},
 \textbf{Shen Li\textsuperscript{2}},
 \textbf{Yong Li\textsuperscript{2}},
 \textbf{Tao Li\textsuperscript{1}}
\\
 \textsuperscript{1}College of Computer Science, Nankai University, Tianjin, China
 \\
 \textsuperscript{2}Alibaba Inc, Hangzhou, China
\\
 \small{
   \textbf{Correspondence:} \href{mailto:email@domain}{litao@nankai.edu.cn}
 }
}

\begin{document}
\maketitle
\begin{abstract}
Large language models achieve exceptional performance on various downstream tasks through supervised fine-tuning. However, the diversity of downstream tasks and practical requirements makes deploying multiple full-parameter fine-tuned models challenging. Current methods that compress the delta weight struggle to achieve ultra-high compression, failing to minimize the deployment overhead. To address the above issue, we propose a novel distribution-driven delta compression framework DeltaDQ, which utilizes Group-wise Dropout and Separate Quantization to achieve ultra-high compression for the delta weight. We have observed that the matrix-computed intermediate results for the delta weight exhibit extremely small variance and min-max range characteristics, referred to as Balanced Intermediate Results. Exploiting this phenomenon, we introduce Group-wise Dropout to perform dropout on the delta weight using an optimal group size. Furthermore, using Separate Quantization, sparse weights are quantized and decomposed to achieve a lower bit. Experimental results show that DeltaDQ achieves $16\times$ compression with improved accuracy compared to baselines for WizardMath and WizardCoder models across different parameter scales. Moreover, DeltaDQ demonstrates the ability for ultra-high compression ratio, achieving $128\times$ compression for the WizardMath-7B model and $512\times$ compression for the WizardMath-70B model.
\end{abstract}

\section{Introduction}
Large Language Models (LLMs) \cite{brown2020language, zhang2022opt, touvron2023llama, achiam2023gpt} have achieved unprecedented advances in recent years, and most researchers and users have adopted the Supervised Fine-Tuning (SFT) \cite{ouyang2022training} to emerge the capabilities of LLMs for a variety of different downstream tasks. SFT enables LLMs to achieve better quality in tasks such as mathematical reasoning and code generation. Meanwhile, despite the existence of many Parameter Efficient Fine-Tuning (PEFT) \cite{ding2023parameter} methods such as LoRA \cite{hu2021lora}, full-parameter fine-tuning models still have higher accuracy under many complex downstream tasks \cite{chen2022revisiting}.

However, the presence of numerous downstream tasks introduces a new challenge: efficiently deploying multiple fine-tuned models during inference. Deploying all fine-tuned models simultaneously would result in high resource consumption and low utilization efficiency. Conversely, loading models on the request demand can cause a sharp increase in latency. While several deployment strategies for multiple LoRA models, such as S-LoRA \cite{sheng2023s} and Punica \cite{chen2023punica}, have been proposed, deploying full-parameter fine-tuned models poses a significant challenge due to their large parameter sizes, which demand substantial memory resources.

\begin{figure}
    \centering
    \includegraphics[width=0.85\linewidth]{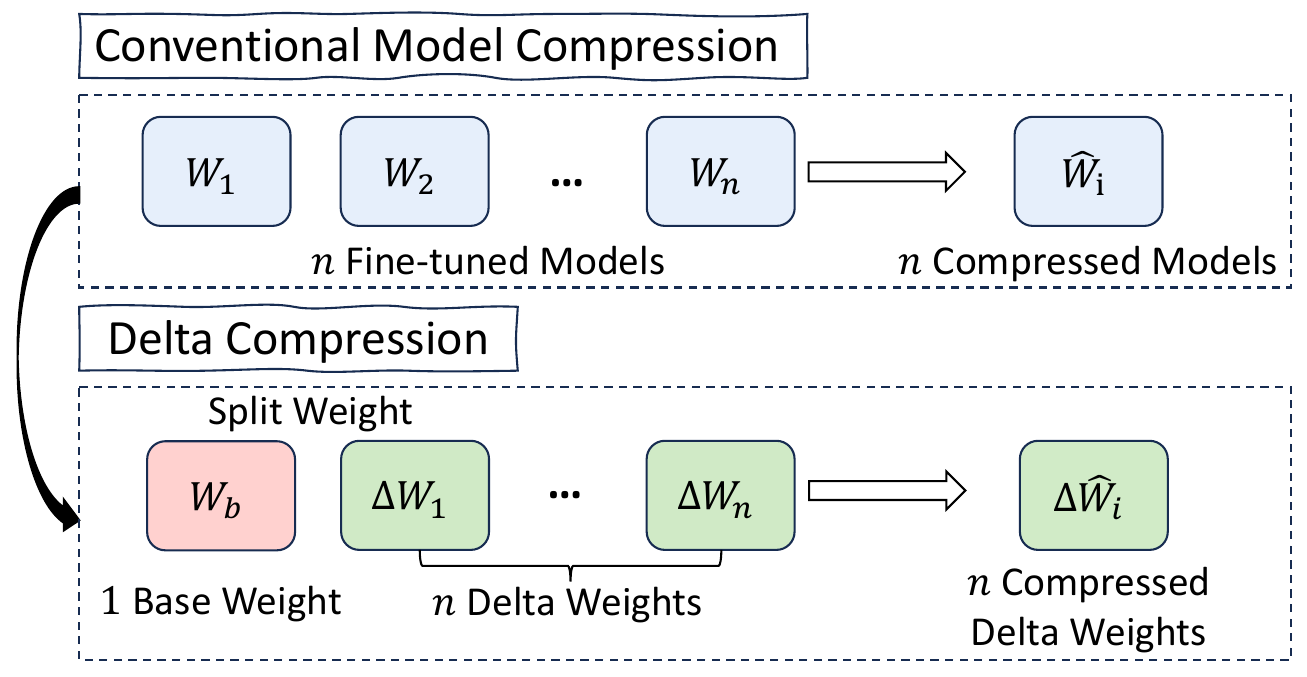}
    \caption{Difference between conventional model compression and delta compression.}
    \label{fig:delta_compression}
\vspace{-10pt}
\end{figure}

Currently, an effective approach to addressing this challenge is through delta compression. As illustrated in Figure \ref{fig:delta_compression}, delta compression reduces overall memory requirements compared to traditional methods by strategically focusing on compressing the more compressible delta weight of each model, rather than the fine-tuned models directly. However, existing methods still fail to achieve a high compression ratio, limiting the number of models that can be deployed on constrained hardware resources. DARE \cite{yu2023language} and BitDelta \cite{liu2024bitdelta}, rely on sole compression techniques like sparsification or quantization, which only achieve a limited compression ratio. Although DELTAZIP \cite{yao2023deltazip} combines sparsity with quantization, it ignores the unique characteristics of delta weight, thereby constraining its overall compression efficacy.

To address the aforementioned challenges, we propose a novel distribution-driven ultra-high delta compression framework called DeltaDQ. We observe that the intermediate results of delta weight during matrix computation exhibit extremely small variance and min-max range distribution, which we refer to as Balanced Intermediate Results. Leveraging this observation, we introduce Group-wise Dropout, which divides the elements of delta weight in each row into groups, then randomly dropout, with the group size $h_g$ optimally determined based on the attention error. Considering the separate computation characteristic in delta compression, we design Separate Quantization, which follows the Group-wise Dropout by quantizing the sparse delta weight to $k$ bits. The quantized weight is then decomposed into $m$ parts based on value, and the quantization bit is reduced to $k - \operatorname{log}m$, effectively minimizing quantization errors even at ultra-low bits. Experimental results show that DeltaDQ achieves nearly lossless $16\times$ compression for WizardMath \cite{luo2023wizardmath} and WizardCoder \cite{luo2023wizardcoder} models of different parameter scales, with significant accuracy improvements compared to baselines. Furthermore, DeltaDQ can compress the WizardMath-7B model by $128\times$ and the WizardMath-70B model by $512\times$, with acceptable accuracy loss.

Our main contributions are as follows:
\begin{itemize}
    \item We discover Balanced Intermediate Results, where the delta weight has better compressibility with extremely small variance and min-max range during matrix computation.
    \item We propose a novel delta compression framework DeltaDQ, which mainly consists of Group-wise Dropout and Separate Quantization to maximize compression ratio.
    \item Experimental results show that DeltaDQ achieves better accuracy than baselines at $16\times$ compression and even higher accuracy than the original model on the WizardCoder model. Meanwhile, DeltaDQ achieves $128\times$ and $512\times$ compression on the WizardMath-7B and WizardMath-70B models, respectively, demonstrating the effectiveness at ultra-high compression.
\end{itemize}
\section{Related Work}
\subsection{Model Compression for LLMs}
Model compression can effectively reduce the deployment costs of LLMs through techniques such as sparsification, quantization, knowledge distillation, and low-rank approximation \cite{tang2024survey, zhou2024survey}. In this paper, we focus on sparsification and quantization. Current quantization methods can primarily be categorized into weight-only and weight-activation quantization. Weight-only quantization, exemplified by methods like GPTQ \cite{frantar2022gptq} and AWQ \cite{lin2023awq}, focuses on reducing model weight precision to lower bits, effectively decreasing memory usage and loading time. Weight-activation quantization, such as LLM.int8() \cite{dettmers2022llm} and SmoothQuant \cite{xiao2023smoothquant}, quantize both weights and activations, further reducing inference memory and latency. Sparsification, represented by pruning methods such as SparseGPT \cite{frantar2023sparsegpt}, Wanda \cite{sun2023simple}, LLM-Pruner \cite{ma2023llm}, and ShearedLLaMA \cite{xia2023sheared}, optimizes model inference by removing unimportant weights. Although these methods work well for LLMs, they do not account for the characteristics of delta weight, which limits their ability to achieve an optimal compression ratio.
\begin{figure*}
    \centering
    \includegraphics[width=\linewidth]{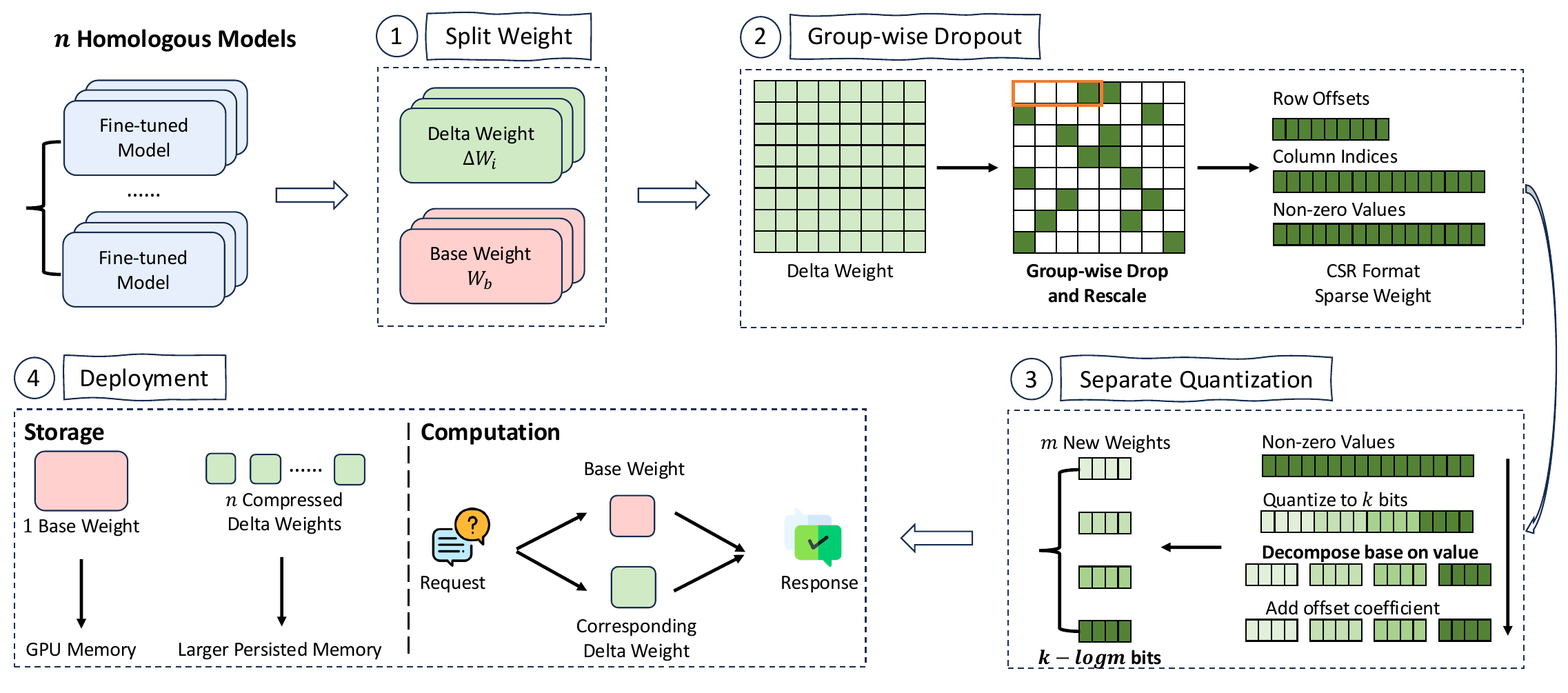}
    \caption{Overview of our delta compression framework DeltaDQ. Our framework is divided into four steps; Step 1: Split Weight; Step 2: Group-wise Dropout; Step 3: Separate Quantization; Step 4: Deployment.}
    \label{fig:overview}
\end{figure*}

\subsection{Multi Fine-Tuned Model Deployment}
The diversity of downstream tasks and user requirements makes deploying multiple fine-tuned models challenging. Current SFT methods are mainly divided into PEFT \cite{han2024parameter} like LoRA \cite{hu2021lora} and QLoRA \cite{dettmers2023qlora}, and full-parameter fine-tuning. For LoRA-based fine-tuned models, frameworks like S-LoRA \cite{sheng2023s} and Punica \cite{chen2023punica} enable the deployment of thousands of LoRA models using efficient parallel strategies and highly optimized CUDA kernels. For full-parameter fine-tuned models, optimization is primarily achieved through delta compression. Although DARE \cite{yu2023language} is designed for model merging, it effectively compresses the delta weight using Dropout \cite{srivastava2014dropout}. BitDelta \cite{liu2024bitdelta} reduces memory requirements by quantizing the delta weight to $1$-bit, while Delta-Come \cite{ping2024delta} applies mixed-precision quantization \cite{wang2019haq} to the delta weight based on singular value magnitudes. DELTAZIP \cite{yao2023deltazip} combines pruning and quantization, building on SparseGPT to compress the delta weight. However, despite being tailored for the delta weight, these methods struggle to achieve beyond $16\times$ compression, which limits the number of models that can be deployed compared to multi-LoRA approaches. Achieving ultra-high compression to increase the efficiency of full-parameter fine-tuned model deployment remains a valuable challenge.

\section{Methodology}

We begin by introducing the foundational concepts of delta compression in Section \ref{subsec:pre}. In Section \ref{subsec:bis}, we describe the Balanced Intermediate Results, highlighting that the delta weight is more easily compressed compared to the fine-tuned weight. Building on this, we propose DeltaDQ, which is primarily composed of the Group-wise Dropout presented in Section \ref{subsec:ard} and the Separate Quantization in Section \ref{subsec:dq}, as shown in Figure \ref{fig:overview}.

\subsection{Preliminaries: Delta Compression}
\label{subsec:pre}
\noindent{\textbf{Algorithm.}} Assume there are $n$ full-parameter fine-tuned models, with their weights denoted as $\{\mathbf{W}_1, \mathbf{W}_2, ..., \mathbf{W}_n\}$, all derived from a homologous base model such as Llama \cite{touvron2023llama}. We can split the weight of each model into two parts, the weight of base model $\mathbf{W}_b$ and the delta weight $\Delta \mathbf{W}_i$ of each fine-tuned model, in the following way:
\begin{equation}
\Delta \mathbf{W}_i = \mathbf{W}_i - \mathbf{W}_b
\enskip.
\end{equation}
The delta compression algorithm is to compress the delta weight $\Delta \mathbf{W}_i$, with the compression ratio specifically applied to the delta weight.

\begin{figure}
    \centering
    \includegraphics[width=0.85\linewidth]{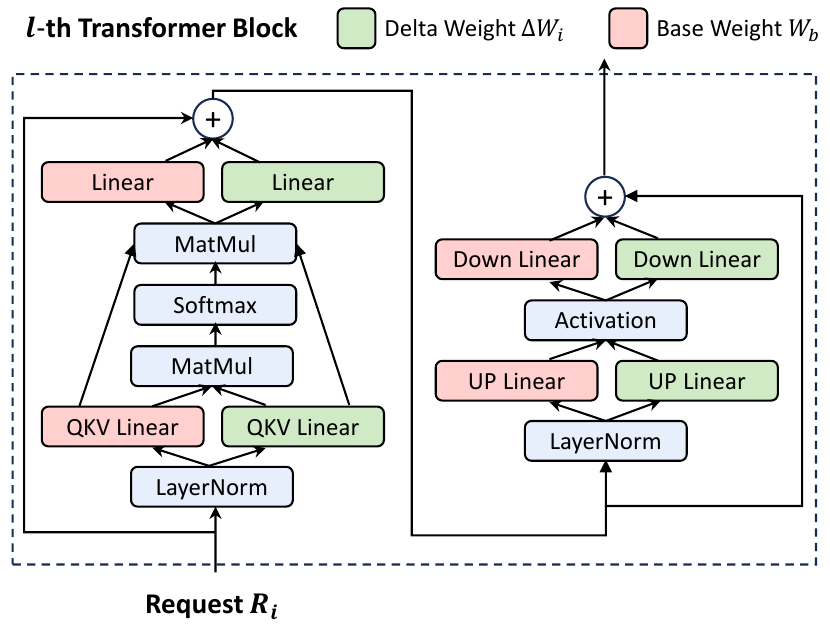}
    \caption{Illustration of Separate Computation. Calculating base weight and delta weight separately in the Linear layer, followed by synchronization.}
    \label{fig:separate_computation}
\vspace{-10pt}
\end{figure}

\noindent{\textbf{Deployment.}} For practical inference deployment, the delta compression framework strategically separates the storage of the base weight $\mathbf{W}_b$ and the compressed delta weight $\Delta \mathbf{W}_i$, as shown in Figure \ref{fig:overview}. In computation, the request $R_i$ for each model follows a separate computation scheme, similar to DELTAZIP \cite{yao2023deltazip}. Specifically, $R_i$ is processed independently by $\mathbf{W}_b$ and $\Delta \mathbf{W}_i$, with synchronization to generate the final outputs, as illustrated in Figure \ref{fig:separate_computation}. Since our method is compatible with other delta compression frameworks at the deployment level, in this paper, we primarily focus on the algorithmic aspects.

\begin{figure*}[htbp] 
  \centering 

  \begin{subfigure}[b]{\textwidth} 
    \centering
    \includegraphics[width=\textwidth]{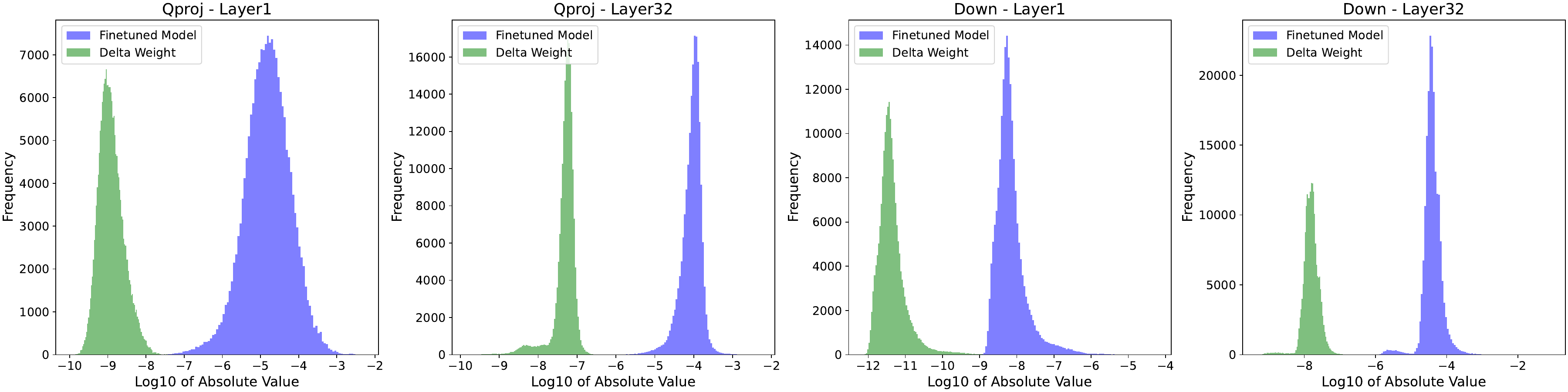} 
    \caption{Variance}
    \label{fig:variance}
  \end{subfigure}\\ 

  \begin{subfigure}[b]{\textwidth}
    \centering
    \includegraphics[width=\textwidth]{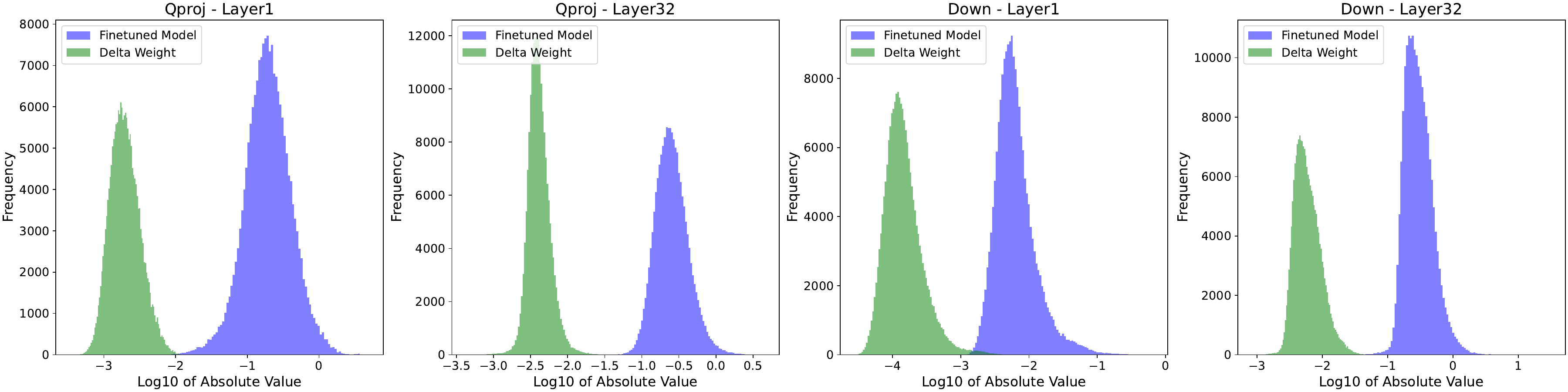} 
    \caption{Min-Max Range}
    \label{fig:range}
  \end{subfigure}

  \caption{Comparison of the variance and min-max range distribution of the intermediate results of the delta weight and fine-tuned weight for each output element matrix computation.} 
  \label{fig:balanced} 
\end{figure*}

\subsection{Balanced Intermediate Results}
\label{subsec:bis}
In model compression, the whole optimization problem is commonly reformulated as a layer-by-layer subproblem. Let $\mathbf{W}_i^l$ represent the weight of the $l$-th layer of model $i$, and $\hat{\mathbf{W}}_i^l$ denote the corresponding compressed weights. The optimization objective of model compression is to minimize the layer-wise $\ell_2$-loss, defined as $\mathcal{L}_{layer}$:
\begin{equation}
\mathcal{L}_{layer} = ||\mathbf{X}_i^l\mathbf{W}_i^{l^T} - \mathbf{X}_i^l\hat{\mathbf{W}}_i^{l^T}||_2^2
\enskip,
\end{equation}
where $\mathbf{X}_i^l$ is the input of the $l$-th layer.

Assuming $\mathbf{X}_i^l \in \mathbb{R}^{t \times h_{in}}$, $\mathbf{W}_i^l,\hat{\mathbf{W}}_i^l \in \mathbb{R}^{h_{out}\times h_{in}}$ and the outputs $\mathbf{A}_i^l,\hat{\mathbf{A}}_i^l \in \mathbb{R}^{t \times h_{out}}$, we can further transform $\mathcal{L}_{layer}$:
\begin{align}
\mathcal{L}_{layer} &= ||\mathbf{X}_i^l\mathbf{W}_i^{l^T} - \mathbf{X}_i^l\hat{\mathbf{W}}_i^{l^T}||_2^2 \\
    &= ||\mathbf{A}_i^l - \hat{\mathbf{A}}_i^l||_2^2  \nonumber \\
    &= \sum_{p=1,q=1}^{t,h_{out}}(a_{p,q}|_i^l-\hat{a}_{p,q}|_i^l)^2 \nonumber
\enskip,
\end{align}
where $a_{p,q}|_i^l$, $\hat{a}_{p,q}|_i^l$ are the elements in the $p$-th row and $q$-th column of outputs $\mathbf{A}_i^l$ and $\hat{\mathbf{A}}_i^l$ before and after compression; for instance, $a_{p,q}|_i^l$ represents:
\begin{equation}
a_{p,q}|_i^l=x_{p,0}w_{q,0}+...+x_{p,h_{in}}w_{q,h_{in}}
\enskip.
\end{equation}

As shown in Figure \ref{fig:balanced}, we find that each element $\Delta a_{p,q}|_i^l$ of the delta weight outputs $\Delta \mathbf{A}_i^l$ has the following two properties compared to the original outputs $\mathbf{A}_i^l$:
\begin{itemize}
    \item Small Variance: The intermediate results $x_{p,0}\Delta w_{q,0},..., x_{p,h_{in}}\Delta w_{q,h_{in}}$ have a small variance between them.
    \item Narrow Min-Max Range: The intermediate results $x_{p,0}\Delta w_{q,0},..., x_{p,h_{in}}\Delta w_{q,h_{in}}$ of $\Delta a_{p,q}|_i^l$ have a very narrow range between the maximum and minimum values.
\end{itemize}
These two properties make delta weight more suitable for compression techniques such as sparsification compared to the original fine-tuned weight. We refer to this phenomenon as Balanced Intermediate Results.

\subsection{Group-wise Dropout}
\label{subsec:ard}
Inspired by the phenomenon of Balanced Intermediate Results, we observe that random dropping methods, such as Dropout, provide a simple yet effective approach to compressing delta weight. In contrast to DARE \cite{yu2023language}, which primarily focuses on model merging, our method leverages the Balanced Intermediate Results by stochasticly dropping weights along matrix computation dimensions (the row dimensions) and subsequently scaling the remaining weights. The procedure for applying Row-wise Dropout to the delta weight $\Delta \mathbf{W}_i^l \in \mathbb{R}^{h_{out}\times h_{in}}$ at the $l$-th layer is as follows:
\begin{itemize}
    \item Row-wise Drop: For a given compression ratio $\alpha$, generate $h_{out}$ random mask vectors $\mathbf{m}_i \in \mathbb{R}^{1 \times h_{in}}$, where $1 - 1 / \alpha$ of the elements in each vector are set to $0$, and the remaining elements are set to $1$. Form a mask matrix $\mathbf{M}$ using these $h_{out}$ mask vectors $\mathbf{m}_i$, and apply it to obtain the masked delta weight as $\Delta \hat{\mathbf{W}}_i^l = \Delta \mathbf{W}_i^l \odot \mathbf{M}$.
    \item Rescaling: Scale the remaining elements of $\Delta \hat{\mathbf{W}}_i^l$ by multiplying $\alpha$, yielding $\Delta \hat{\mathbf{W}}_i^l = \alpha \cdot \Delta \hat{\mathbf{W}}_i^l$.
\end{itemize}
Through the aforementioned steps, we obtain the sparse $\Delta \hat{\mathbf{W}}_i^l$. For any given element $\Delta a_{p,q}|_i^l$, employing Row-wise Dropout, as opposed to global Dropout like DARE, more effectively minimizes the compression-induced error $\Delta a_{p,q}|_i^l-\Delta \hat{a}_{p,q}|_i^l$, thereby approximating $\mathcal{L}_{layer}=0$.

\begin{figure}
    \centering
    \includegraphics[width=0.7\linewidth]{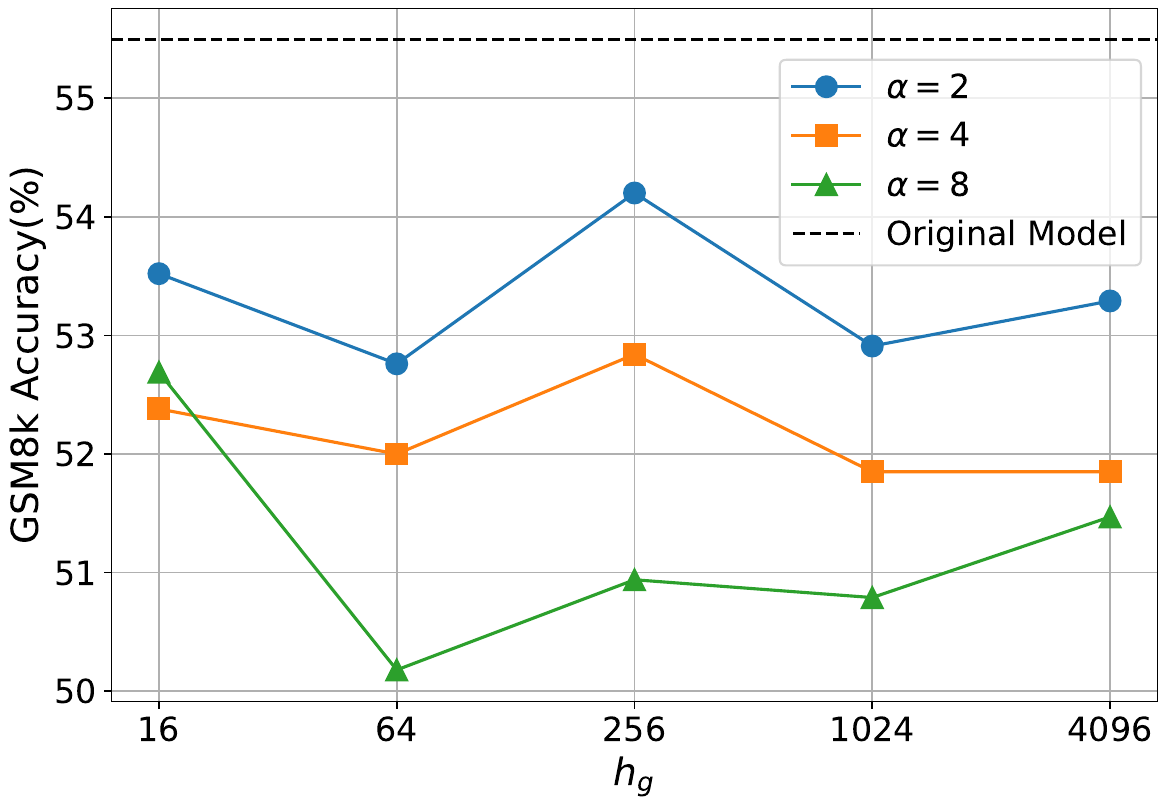}
    \caption{Impact of different group sizes on the accuracy of WizardMath-7B at the same compression ratio.}
    \label{fig:dropout}
\vspace{-10pt}
\end{figure}

Grouping techniques are widely employed in model compression, often significantly improving the accuracy. As illustrated in Figure \ref{fig:dropout}, under a fixed compression ratio $\alpha$, subdividing $\mathbf{m}_i$ into $h_{in} / {h_g}$ finer-grained mask groups $\mathbf{m}_{i,j} \in \mathbb{R}^{1 \times h_g} (h_g \leq h_{in})$ where $h_g$ represents the group size, affects model accuracy. Unlike group-wise quantization, a smaller $h_g$ does not necessarily result in higher accuracy and the optimal group size $h_g^*$ varies across different models. Nonetheless, compared to Row-wise Dropout, the grouping method often achieves superior results at the optimal group size $h_g^*$. To further mitigate accuracy loss, we extend our approach to Group-wise Dropout.

To efficiently determine the optimal group size $h_g^*$, we first constrain the weights across all layers and each row within these weights to use the same group size. We define the range of $h_g$ as $\{\alpha, \alpha*2^1, \alpha*2^2, ..., h_{in}\}$. Directly selecting $h_g^*$ based on task accuracy is resource-intensive, so we employ an efficient proxy metric for this selection. In Transformer \cite{vaswani2017attention}, the attention matrix, which captures the relationships between different tokens in a sequence, is a crucial feature that must maintain minimal loss after compression. Moreover, the shallow layers of a model are generally more sensitive to compression than the deeper layers \cite{yin2023outlier}. Based on these, we use the error in the attention matrix before and after compression as a proxy metric for evaluating group size:
\begin{equation}
\mathcal{E}_p = || \mathbf{Q}_1\mathbf{K}_1^T - \hat{\mathbf{Q}}_1\hat{\mathbf{K}}_1^T ||_2^2
\enskip,
\end{equation}
where $\mathbf{Q}_1$ and $\mathbf{K}_1$ represent the query and key of the $1$-th layer, respectively, then $\hat{\mathbf{Q}}_1$ and $\hat{\mathbf{K}}_1^T$ are their compressed counterparts. We select the group size $h_g$ with the smallest error as the optimal group size for the model. Through this approach, computations in the subsequent layers are avoided, effectively reducing the search time. Additionally, we use only $1\%$ of the original test data as the evaluation dataset, further decreasing the search time.

\begin{table*}
  \centering
  \begin{tabular}{ccccccccc}
    \toprule
    \multirow{2}{*}{\textbf{Method}} & \multirow{2}{*}{\textbf{Quantization}} & \multirow{2}{*}{\textbf{\makecell{Compression \\ Ratio $\alpha$}}} & \multicolumn{3}{c}{\textbf{WizardMath}} & \multicolumn{3}{c}{\textbf{WizardCoder}} \\
    \cmidrule(lr){4-6} \cmidrule(lr){7-9}
    & & & \textbf{7B} & \textbf{13B} & \textbf{70B} & \textbf{7B} & \textbf{13B} & \textbf{34B} \\
    \midrule
    Original & \ding{55} & 1 & 55.49 & 63.83 & 81.80 & 55.48 & 64.02 & 73.17 \\
    \midrule
    Magnitude & \ding{55} & 2 & 52.00 & \textbf{65.04} & 74.29 & 46.95 & 63.41 & 69.51 \\
    DELTAZIP & \ding{55} & 2 & 53.60 & 64.59 & \textbf{81.65} & 53.05 & 62.80 & 71.95 \\
    DARE & \ding{55} & 2 & 53.67 & 64.36 & 80.89 & 57.31 & 62.80 & 73.17\\
    DeltaDQ & \ding{55} & 2 & \textbf{54.20} & 63.92 & \textbf{81.65} & \textbf{58.32} & \textbf{65.24} & \textbf{75.00} \\
    \midrule
    Magnitude & \ding{55} & 4 & 45.71 & 61.71 & 44.20 & 10.36 & 12.19 & 35.36 \\
    DELTAZIP & \ding{55} & 4 & 50.87 & 62.17 & \textbf{81.96} & 55.49 & 64.63 & 68.29 \\
    DARE & \ding{55} & 4 & 51.70 & 63.52 & 80.21 & \textbf{57.92} & 61.58 & 73.17 \\
    DeltaDQ & \ding{55} & 4 & \textbf{52.84} & \textbf{64.51} & 81.95 & 55.48 & \textbf{65.24} & \textbf{74.39} \\
    \midrule
    Magnitude & \ding{55} & 8 & 32.37 & 55.99 & 30.47 & 6.70 & 2.43 & 0.00 \\
    DELTAZIP & \ding{55} & 8 & 46.63 & 59.29 & 80.82 & 42.68 & 59.15 & 69.51 \\
    DARE & \ding{55} & 8 & 50.11 & 63.07 & 80.28 & \textbf{56.70} & 62.19 & 71.95 \\
    DeltaDQ & \ding{55} & 8 & \textbf{52.69} & \textbf{64.13} & \textbf{81.50} & \textbf{56.70} & \textbf{65.24} & \textbf{72.56} \\
    \midrule
    Magnitude & \ding{55} & 16 & 15.84 & 39.72 & 38.43 & 0.60 & 0.00 & 3.04 \\
    DELTAZIP & \ding{51} & 16 & 47.38 & 59.13 & 80.74 & 38.41 & 52.46 & 68.29 \\
    DARE & \ding{55} & 16 & 48.59 & 61.78 & 81.12 & 57.31 & 64.02 & 71.95  \\
    DeltaDQ & \ding{51} & 16 & \textbf{52.99} & \textbf{63.98} & \textbf{81.57} & \textbf{58.53} & \textbf{65.24} & \textbf{73.17} \\
    \bottomrule
  \end{tabular}
  \caption{Accuracy comparison of WizardMath and WizardCoder at various compression ratios, with bold highlighting to indicate the top-performing method for each case. "Quantization" indicates whether quantization is utilized.}
  \label{tab:basic_compression}
\end{table*}

\begin{figure}
    \centering
    \includegraphics[width=\linewidth]{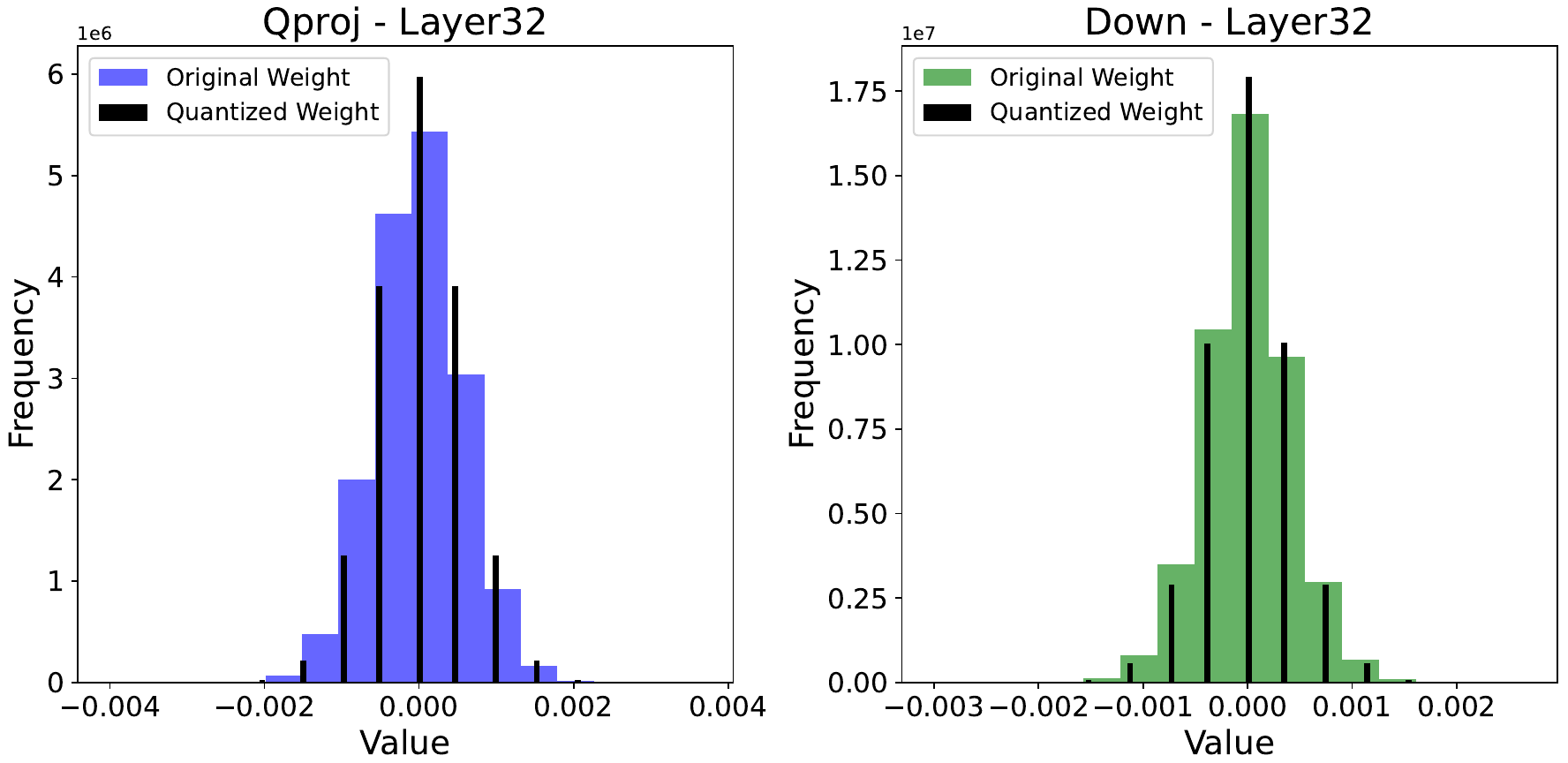}
    \caption{The distribution of delta weight before and after uniform quantization.}
    \label{fig:distribution}
\vspace{-10pt}
\end{figure}

\subsection{Separate Quantization}
\label{subsec:dq}

Despite Group-wise Dropout achieving over $10\times$ compression on the delta weight in most fine-tuned models, further increasing the compression ratio $\alpha$ often leads to significant accuracy loss. As shown in Figure \ref{fig:distribution}, the distribution of delta weight demonstrates strong compatibility with uniform quantization. However, the primary limitation of quantization is the substantial accuracy degradation at ultra-low bits, which also occurs with the delta weight. The challenge remains in reducing the quantization bit further to enhance compression ratio without incurring accuracy loss.

Since we are quantizing weight $\Delta \hat{\mathbf{W}}_i^l$, following a weight-only scheme, offline quantization is feasible. Additionally, when splitting a fine-tuned model into base weight and delta weight, as discussed in Section \ref{subsec:pre}, there is an inherent separate computation during deployment. And the weight $\Delta \hat{\mathbf{W}}_i^l$ is a sparse matrix stored in Compressed Sparse Row (CSR) format, where row offsets, column indices, and non-zero values are recorded. When a sparse matrix is decomposed into $m$ parts, the storage overhead for column indices and non-zero values remains unchanged, while the row offsets increase by $m-1$ times, which is a negligible increase. Furthermore, each decomposed weight becomes even sparser, which is beneficial for computations using sparse libraries \cite{naumov2010cusparse}. These characteristics allow us to apply separate quantization on the sparsified delta weight. We can directly decompose the delta weight based on its magnitude, thereby reducing the quantization bit without compromising accuracy.

As illustrated in Figure \ref{fig:overview}, given a compression ratio of $\alpha_1$ from the previous step, we first quantize $\Delta \hat{\mathbf{W}}_i^l$ using the following per-tensor granularity uniform quantizer:
\begin{equation}
\mathcal{Q}_{i}^l = \operatorname{clip}(\lfloor \frac{\Delta \hat{\mathbf{W}}_{i}^l}{s} \rceil + z, 0, 2^{k} - 1)
\enskip,
\end{equation}
\begin{equation}
\quad s = \frac{\max(\Delta \hat{\mathbf{W}}_{i}^l) - \min(\Delta \hat{\mathbf{W}}_{i}^l)}{2^{k}-1}
\enskip,
\end{equation}
\begin{equation}
z = \lfloor \frac{-\min(\Delta \hat{\mathbf{W}}_{i}^l)}{s} \rceil
\enskip,
\end{equation}
where $\mathcal{Q}_{i}^l$ represents the quantized weights, $k$ denotes the quantization bit, $s$ and $z$ represent the scale factor and zero point, respectively. After quantization, the elements in $\mathcal{Q}_{i}^l$ belong to the set $\{0, 1, ..., 2^{k}-1\}$. We then decompose $\mathcal{Q}_{i}^l$ into $m$ new quantized weights $\mathcal{Q}_{i,j}^l$ based on the value:
\begin{equation}
\mathcal{Q}_{i,j}^l = \mathcal{Q}_{i}^l \cdot \mathbb{I}(r^{min}_j \leq \mathcal{Q}_{i}^l \leq r^{max}_j) + o_j
\enskip,
\end{equation}
\begin{equation}
r^{min}_j = \frac{2^k}{m} * (j - 1), r^{max}_j = \frac{2^k}{m} * j - 1
\enskip,
\end{equation}
\begin{equation}
o_j = - \frac{2^k}{m} * (j - 1) \quad (j = 1, ..., m)
\enskip.
\end{equation}
Here, $\mathbb{I}(\cdot)$ represents the elements that are filtered within the range $[r^{min}_j, r^{max}_j]$, and $o_j$ represents the offset coefficient. The corresponding dequantization process is:
\begin{equation}
\mathcal{DQ}_{i,j}^l = s \cdot (\mathcal{Q}_{i,j}^l - z - o_j)
\enskip.
\end{equation}
In this way, we can reduce the storage bit of each new weight $\mathcal{Q}_{i,j}^l$ to $k - \operatorname{log}m$ bits, allowing us to achieve $2$-bit or even $1$-bit quantization, thereby increasing the compression ratio to $\alpha_1 * 16 / (k - \operatorname{log}m)$.

\begin{table}[h]
  \centering
  \begin{tabular}{ccc}
    \toprule
    \textbf{Method} & \textbf{\makecell{Compression \\ Ratio $\alpha$}}  & \textbf{\makecell{Wizard \\ Math-7B}} \\
    \midrule
    Original  &  1 & 55.49 \\
    \midrule
    Magnitude &  32 & 2.27 \\
    DELTAZIP &  32 & 46.47\\
    DARE &  32 & 46.09 \\
    DeltaDQ($m=1$) &  32 & \textbf{52.69}\\
    \midrule
    Magnitude &  64 & 0.30 \\
    DELTAZIP &  64 & 45.94 \\
    DARE &   64 & 29.94 \\
    DeltaDQ($m=1$) & 64 & 33.43 \\
    DeltaDQ($m=4$) &  64 & \textbf{52.69} \\
    \midrule
    Magnitude &  128 & 0.00 \\
    DELTAZIP &  128 & 26.61 \\
    DARE &   128 & 1.81 \\
    DeltaDQ($m=1$) & 128 & 0.00  \\
    DeltaDQ($m=8$) &  128 & \textbf{52.69} \\
    \midrule
    DeltaDQ($m=16$) &  - & \textbf{52.69} \\
    \bottomrule
  \end{tabular}
  \caption{Accuracy comparison of WizardMath-7B under ultra-high compression ratio. "-" indicates extreme compression, where all values in each new weight are identical, requiring only one value to be stored.}
  \label{tab:ultra_7b}
\end{table}

\section{Experiment}

\subsection{Setup}
\noindent{\textbf{Baseline.}} We compare DeltaDQ with three representative compression methods. Magnitude \cite{han2015learning} is a classical pruning approach based on weight magnitudes and serves as a strong baseline. Additionally, we include two delta compression techniques: DELTAZIP \cite{yao2023deltazip}, which employs both sparsification and quantization similar to our method, and DARE \cite{yu2023language}, which relies solely on sparsification.

\begin{table}[h]
  \centering
  \begin{tabular}{ccc}
    \toprule
    \textbf{Method} & \textbf{\makecell{Compression \\ Ratio $\alpha$}}  & \textbf{\makecell{Wizard \\ Math-70B}} \\
    \midrule
    Original  &  1 & 81.80 \\
    \midrule
    Magnitude &  128 & 0.98 \\
    DELTAZIP &  128 & 73.91 \\
    DARE &  128 & 79.07 \\
    DeltaDQ($m=1$) &  128 & \textbf{79.90} \\
    \midrule
    Magnitude &  256 & 0.07 \\
    DELTAZIP &  256 & 73.61 \\
    DARE &   256 & 71.72 \\
    DeltaDQ($m=1$) & 256 & 14.25 \\
    DeltaDQ($m=4$) &  256 & \textbf{79.90} \\
    \midrule
    Magnitude &  512 & 0.00 \\
    DELTAZIP &  512 & 48.74 \\
    DARE &   512 & 37.45 \\
    DeltaDQ($m=1$) & 512 & 0.00 \\
    DeltaDQ($m=8$) &  512 & \textbf{79.90} \\
    \midrule
    DeltaDQ($m=16$) & - & \textbf{79.90} \\
    \bottomrule
  \end{tabular}
  \caption{Accuracy comparison of WizardMath-70B under ultra-high compression ratio.}
  \label{tab:ultra_70b}
\end{table}

\noindent{\textbf{Models, Datasets and Evaluation.}} We evaluate three types of fine-tuned models at different parameter scales: WizardMath \cite{luo2023wizardmath}, WizardLM \cite{xu2023wizardlm}, and WizardCoder \cite{luo2023wizardcoder}, where the first two are fine-tuned from Llama2, and the latter is fine-tuned from CodeLlama \cite{roziere2023code}. Our evaluation primarily uses two datasets: GSM8k \cite{cobbe2021training} for assessing WizardMath, and HumanEval \cite{chen2021evaluating} for WizardCoder.

\noindent{\textbf{Implementation Details.}} DeltaDQ is built with PyTorch \cite{paszke2017automatic} and utilizes models and datasets from Huggingface Transformers \cite{wolf2019huggingface}, with accuracy assessments conducted through the vLLM framework \cite{kwon2023efficient}. Other methods use open-source implementations for evaluation. All our experiments are conducted on $8$ NVIDIA V100 GPUs with $32$G of memory and $8$ NVIDIA A100 GPUs with $80$G of memory.

\subsection{Main Results}
\noindent{\textbf{Basic Compression.}} We first compare the accuracy against baselines at $2\times$, $4\times$, $8\times$, and $16\times$ basic compression ratio. As shown in Table \ref{tab:basic_compression}, our framework achieves better accuracy for WizardMath and WizardCoder models compared to the remaining three methods in most cases. At $2\times$, $4\times$, and $8\times$ compression ratio, DeltaDQ only utilizes Group-wise Dropout, yet it still outperforms the baselines, particularly at a higher compression ratio. And we observe that for a lower compression ratio, simpler methods can also achieve strong results. For example, at $2\times$ compression on the WizardMath-13B model, the Magnitude method delivers the best performance. At a $16\times$ compression ratio, our framework surpasses the state-of-the-art accuracy for the WizardMath-7B and 13B models by $4.40$ and $2.20$, respectively, and outperforms the original WizardCoder-7B and 13B models by $3.05$ and $1.22$. Additionally, we find that models with larger parameter sizes tend to be easier to compress.

\begin{table}[t]
    \centering
    \begin{tabular}{cccc}
    \toprule
    \textbf{$\alpha$} & \textbf{Selection Method} & \textbf{Time(minutes)} & \textbf{$h_g^*$} \\
    \midrule
    \multirow{2}{*}{2} & Direct & 651 & 256 \\
    & Proxy & \textbf{217} & 256 \\
    \midrule
    \multirow{2}{*}{4} & Direct & 590 & 256 \\
    & Proxy & \textbf{193} & 256 \\
    \midrule
    \multirow{2}{*}{8} & Direct & 533 & 16 \\
    & Proxy & \textbf{168}& 16 \\
    \bottomrule
    \end{tabular}
    \caption{Comparison of effects of different group size selection methods on the WizardMath-7B model.}
    \label{tab:select}
\end{table}

\noindent{\textbf{Ultra-High Compression.}} As shown in Table \ref{tab:ultra_7b} and \ref{tab:ultra_70b}, we achieve outstanding results at ultra-high compression ratio. With $32\times$ compression on the WizardMath-7B model and $128\times$ compression on the WizardMath-70B model, DeltaDQ improves accuracy over baselines by $6.60$ and $0.83$, respectively. At ultra-high compression, DeltaDQ leverages Separate Quantization to increase the compression ratio by enlarging $m$ without sacrificing accuracy. For example, with $128\times$ compression of the WizardMath-7B model, we first achieve $8\times$ compression using Group-wise Dropout. Then, through Separate Quantization, we decompose the compressed weights into $8$ parts and quantize each part to $1$-bit. For the $512\times$ compression of the WizardMath-70B model, we first achieve $32\times$ compression using Group-wise Dropout, followed by Separate Quantization. This leads to a significant improvement in accuracy compared to baselines.

\subsection{Analysis}
\noindent{\textbf{Evaluation of Group-wise Dropout.}} In Group-wise Dropout, achieving optimal accuracy requires searching for the optimal group size $h_g^*$. As shown in Table \ref{tab:select}, our proposed proxy metric significantly optimizes the search speed compared to directly testing accuracy. Under three different $\alpha$ settings, the search time is reduced to about 30\% of the Direct method while still achieving the same results.

\begin{figure}[h]
    \centering
    \includegraphics[width=\linewidth]{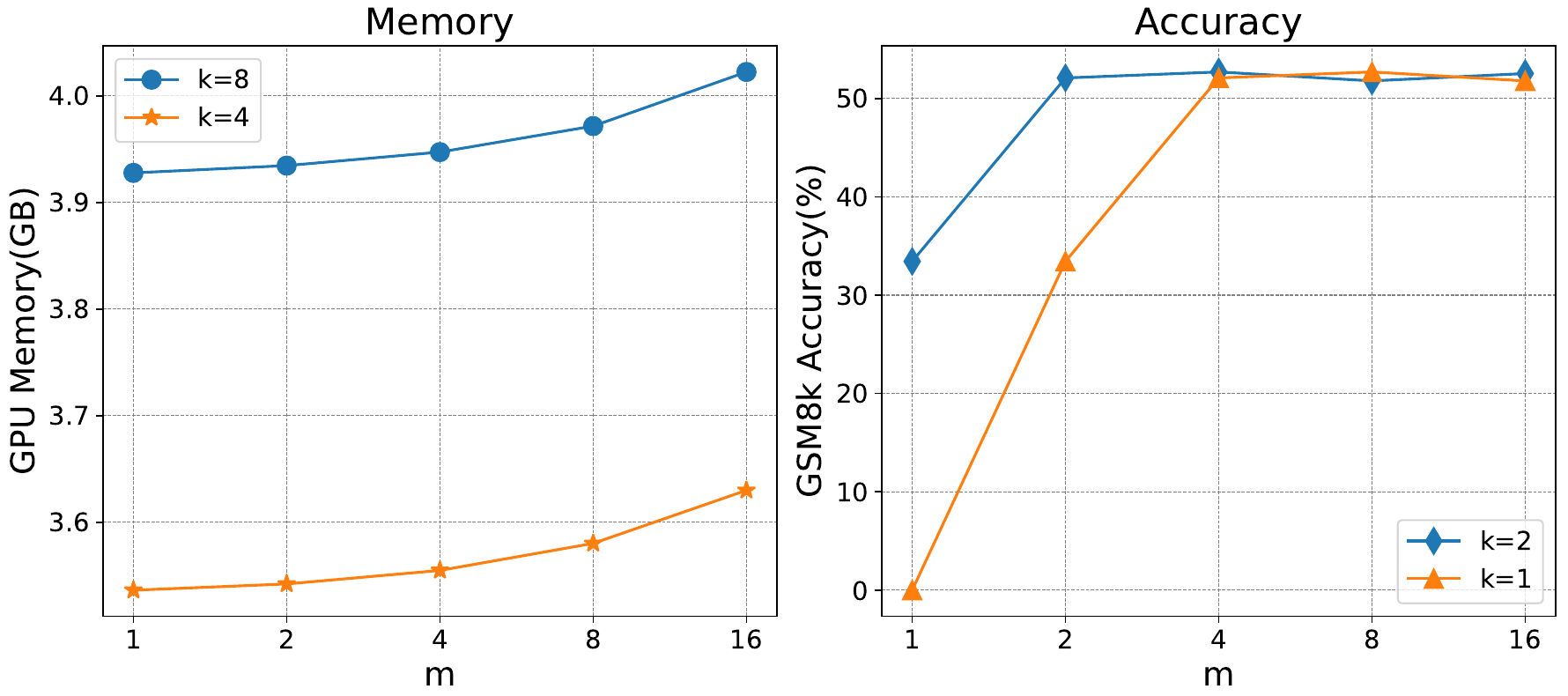}
    \caption{The impact of Separate Quantization on memory and accuracy in the WizardMath-7B model. Here, $k$ represents the final quantization bit of the delta weight.}
    \label{fig:s_effect}
\end{figure}

\begin{figure}[h]
    \centering
    \includegraphics[width=\linewidth]{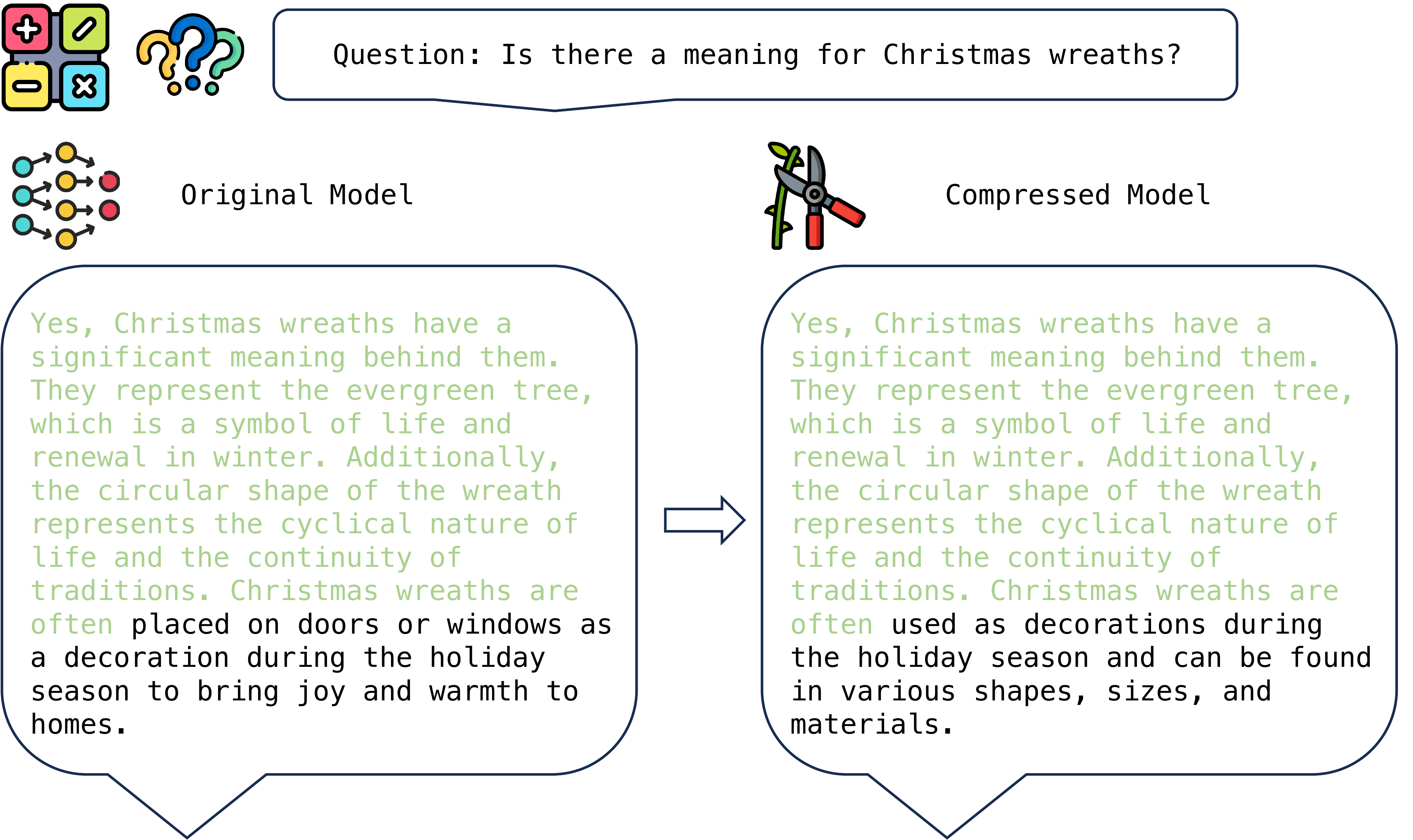}
    \caption{Comparing responses of the WizardLM-7B
model before and after $128\times$ compression.}
    \label{fig:case_study}
\end{figure}

\noindent{\textbf{Evaluation of Separate Quantization.}} For Separate Quantization, as shown in Figure \ref{fig:s_effect}, we analyze the impact of decomposing on GPU memory and accuracy. When quantized to $8$-bit and $4$-bit, the memory footprint of the model parameters remains nearly unchanged as the number of parts increases. This is because, after decomposing, the sparse matrix storage only adds the row offsets and quantization parameters offset coefficient, both of which require negligible memory. When quantized to $2$-bit and $1$-bit, the accuracy of model increases notably as $m$ grows. Although each part still represents values with ultra-low bits, the overall representational capacity improves with more parts, leading to better performance.

\noindent{\textbf{Case Study.}} We additionally evaluate the change in response of WizardLM-7B before and after DeltaDQ. As seen in Figure \ref{fig:case_study}, the responses generated by the model before and after applying DeltaDQ exhibit a high degree of similarity for the identical question at $128\times$ compression. This illustrates the generalization of our framework to different types of fine-tuned models and its non-awareness to practical users.
\section{Conclusion}
In this paper, we introduce DeltaDQ, a novel delta compression framework primarily composed of Group-wise Dropout and Separate Quantization. Group-wise Dropout leverages the inherent properties of delta weight to dropout in a group-wise manner, while Separate Quantization applies further compression to the sparse delta weight. Experimental results demonstrate that DeltaDQ can effectively achieve ultra-high delta compression.

\section*{Limitations}
While our framework shows significant potential, its deployment performance is currently hindered by the absence of optimized libraries for accelerating operations with low-bit sparse weights. Such specialized libraries are essential to fully realize the efficiency gains of our approach, as they would enable faster computation and more efficient memory access patterns tailored to the weight structure. This is also part of our future work.

\section*{Ethical Considerations}
Our framework may introduce slight variations in model outputs, as the process alters the precise values of the weights, potentially affecting inference responses. However, it is designed to minimize these effects, prioritizing the preservation of performance quality and output consistency.

\bibliography{ref}

\end{document}